\documentclass[sigconf,nonacm]{acmart} 

\usepackage{tikz}
\usetikzlibrary{shapes.geometric, arrows.meta, positioning, calc, backgrounds, fit, shadows}
\usepackage{booktabs}
\usepackage{tabularx}
\usepackage{ragged2e}
\usepackage{multirow}

\setcopyright{none}              
\renewcommand\footnotetextcopyrightpermission[1]{} 

\acmConference[]{}{}{}
\acmBooktitle{}

\definecolor{darkText}{RGB}{50, 50, 60}

\begin{document}

\title{Generative AI for Video Trailer Synthesis: \\ From Extractive Heuristics to Autoregressive Creativity}
\thanks{Accepted at the WSDM 2026 Workshop on Generative AI for Streaming Media and the IEEE WACV 2026 Workshop on Video Analysis for Long-form Content and Event Detection (VALED).}

\author{Abhishek Dharmaratnakar}
\affiliation{%
  \institution{Google LLC}
  \city{San Bruno}
  \country{USA}
}

\author{Srivaths Ranganathan}
\affiliation{%
  \institution{Google LLC}
  \city{Mountain View}
  \country{USA}
}

\author{Debanshu Das}
\affiliation{%
  \institution{Google LLC}
  \city{Mountain View}
  \country{USA}
}

\author{Anushree Sinha}
\affiliation{%
  \institution{Google LLC}
  \city{Mountain View}
  \country{USA}
}

\begin{abstract}
The domain of automatic video trailer generation is currently undergoing a profound paradigm shift, transitioning from heuristic-based extraction methods to deep generative synthesis. While early methodologies relied heavily on low-level feature engineering, visual saliency, and rule-based heuristics to select representative shots, recent advancements in Large Language Models (LLMs), Multimodal Large Language Models (MLLMs), and diffusion-based video synthesis have enabled systems that not only identify key moments but also construct coherent, emotionally resonant narratives. This survey provides a comprehensive technical review of this evolution, with a specific focus on generative techniques including autoregressive Transformers, LLM-orchestrated pipelines, and text-to-video foundation models like OpenAI's Sora and Google's Veo. We analyze the architectural progression from Graph Convolutional Networks (GCNs) to Trailer Generation Transformers (TGT), evaluate the economic implications of automated content velocity on User-Generated Content (UGC) platforms, and discuss the ethical challenges posed by high-fidelity neural synthesis. By synthesizing insights from recent literature, this report establishes a new taxonomy for AI-driven trailer generation in the era of foundation models, suggesting that future promotional video systems will move beyond extractive selection toward controllable generative editing and semantic reconstruction of trailers.
\end{abstract}



\maketitle


\section{Introduction}

The proliferation of digital video platforms, ranging from streaming services hosting feature films to User-Generated Content (UGC) platforms like YouTube and TikTok, has created an unprecedented surplus of video data [13]. In this hyper-saturated ecosystem, the movie trailer or its short-form equivalent, the "teaser" or "short" serves as the primary vector for viewer conversion and engagement. Trailer creation is a traditionally labor-intensive artistic pursuit, requiring human editors to meticulously condense extensive footage into a gripping narrative that effectively balances the setup, rising action, and climax, without compromising key plot details [36]. The economic pressure to produce these assets at scale, particularly for the long tail of UGC content, has driven intense research into automation.

First-generation systems employed heuristic rules based on motion intensity, audio energy, and shot boundary detection to perform extractive summarization [17]. Second-generation approaches integrated supervised deep learning, utilizing Convolutional Neural Networks (CNNs) [10] and Recurrent Neural Networks (RNNs) to classify "trailer-worthy" shots based on learned features. While these discriminative models were more accurate in specific tasks (improving precision), they often lacked the ability to comprehend the complete narrative structure essential for creating a truly compelling trailer.

However, we are currently witnessing a third wave: the \textbf{Generative Shift}. This phase is characterized by the move from discriminatory tasks (classification/ranking) to generative tasks (synthesis/reconstruction). Current state-of-the-art systems do not merely select shots; they employ Large Language Models (LLMs) to script voice-overs, synthesize background music aligned with emotional beats, and utilize autoregressive Transformers to predict optimal shot sequencing rather than relying on chronological order [8]. Furthermore, the emergence of text-to-video (T2V) foundation models such as Sora and Veo 2 suggests a future where trailers may be synthesized from textual prompts with visually compelling and often physically plausible sequences, partially decoupling trailer creation from the constraints of existing footage [33].

This report surveys this generative transition, analyzing the architectures, datasets, and evaluation metrics that define modern ATG. Unlike previous surveys that focus on general video summarization, we concentrate specifically on the persuasive and narrative requirements of trailers, examining how Generative AI (GenAI) addresses the complex interplay of visual attractiveness, auditory rhythm, and semantic coherence.

\section{Problem Formulation and The Generative Shift}

\subsection{Distinguishing Trailers from Summaries}

It is critical to rigorously distinguish trailer generation from the broader, more established field of video summarization. While these domains share technological roots shot boundary detection, keyframe extraction, and multimodal analysis, their objectives are fundamentally orthogonal, and often contradictory. Summarization aims to maximize information coverage and minimize redundancy to provide a synopsis; it is a utility-driven task intended to save the viewer time by revealing the plot. In stark contrast, trailer generation is an adversarial and persuasive task. A trailer must induce curiosity (often by withholding information), evoke specific emotional states (fear, excitement, melancholy), and adhere to cinematic pacing rules that may contradict information maximization [17].

Formally, let a video $V$ be a sequence of shots $\{s_1, s_2,..., s_N\}$.
Summarization seeks a subset $S \subset V$ such that the semantic distance between $S$ and $V$ is minimized, effectively compressing the narrative. Ideally, $S$ allows the viewer to understand the full story without watching $V$.
Trailer Generation seeks a subset (or synthesized sequence) $T$ that maximizes a predicted engagement score $E(T)$ or attractiveness function $A(T)$, subject to temporal constraints $L_{min} \le |T| \le L_{max}$ and narrative ordering constraints [9]. Crucially, $T$ must not allow the viewer to fully understand the resolution of $V$.
This distinction necessitates different loss functions and architectural choices. Summarization models often penalize redundancy heavily; however, a trailer might intentionally repeat a visual motif (e.g., a ticking clock or a recurring villain) to build tension. Summarization models preserve chronological order to maintain causality; trailer generation models often reorder shots (non-chronological sequencing) to create juxtapositions that imply relationships or heighten drama. The shift to generative models allows us to optimize for these persuasive metrics directly, rather than using proxies like "representativeness."

\begin{figure}[t]
    \centering
    \includegraphics[width=0.85\linewidth, trim={0 7cm 0 7cm}, clip]{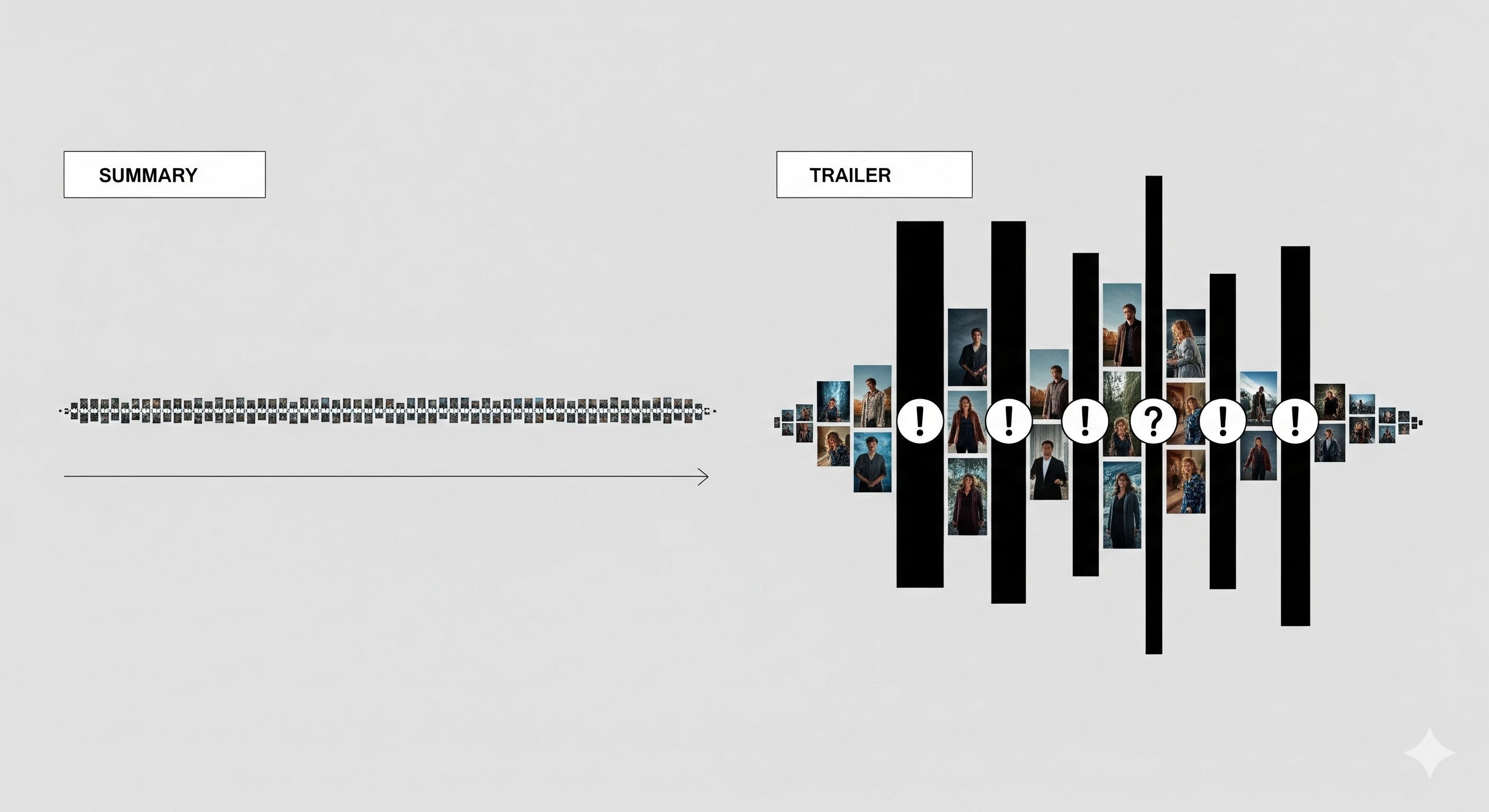}
    \Description{A diagram contrasting video summarization which shrinks content, versus video trailer generation which highlights emotional beats.}
    \caption{Video Summary v/s Video Trailer}
    \label{fig:placeholder}
\end{figure}

\subsection{The Taxonomy of Trailer Generation}

To structure our analysis of the field's evolution, we propose a taxonomy that categorizes ATG approaches based on their reliance on generative capabilities versus extractive logic.

\begin{table}[t]
\caption{Taxonomy of Automatic Trailer Generation}
\label{tab:taxonomy}
\small
\renewcommand{\arraystretch}{0.9}
\setlength{\tabcolsep}{3pt}
\centering
\begin{tabularx}{\linewidth}{l l l X}
\toprule
\textbf{Category} & \textbf{Mechanism} & \textbf{Modalities} & \textbf{Key Techniques} \\
\midrule
\textbf{Heuristic} & Rule-based & Motion, Audio & Shot Boundary, Clustering [17] \\
\textbf{Affective} & Saliency & Gaze, Emotion & Point Processes [37], SVMs \\
\textbf{Graph-Based} & Structural & Visual Relations & GCNs [3], Story Beats \\
\midrule
\textbf{LLM-Agent} & Planning & Text + Visual & GPT-4, CLIP, TTS [8] \\
\textbf{Autoregressive} & Translation & Visual Tokens & Transformers (TGT) [9] \\
\textbf{Foundation} & Diffusion & Text-to-Video & Sora, Veo 2 [33] \\
\bottomrule
\end{tabularx}
\end{table}

\section{Pre-Generative Foundations}

The earlier approaches highlighted the importance of quantifying a shot's 'attractiveness' and its relation to surrounding clips, principles that are now internalized within the underlying architectures of modern Transformers and diffusion models. Extensive research has explored extracting keyframes based on genre detection [11], unsupervised anomaly detection [1], and sport-specific summarization [28]. Similar techniques have been applied to smart subtitle-based extraction [35] and comprehensive skimming taxonomies [23, 26, 20].

\subsection{Affective and Point Process Models}

Early attempts to quantify "trailer-worthiness" moved beyond simple signal processing (like motion detection) to focus on biological and psychological proxies for engagement. A seminal contribution in this area came from Xu et al. (2015), who introduced the concept of \textbf{fixation variance} as a surrogate measure for visual attractiveness [37]. The underlying hypothesis posits that attractive scenes (e.g., intense action, emotional close-ups, recognizable character faces) cause viewer gazes to converge on specific focal points (low variance), while boring or cluttered scenes cause gazes to wander across the screen (high variance).

To utilize this physiological data for generation, they modeled the dynamics of attractiveness using a \textbf{Self-Correcting Point Process}. This mathematical framework assumes that a viewer's need for stimulation increases over time; if the video remains static or boring, the "intensity" or demand for a new event rises. The intensity function $\lambda(t)$ of the process increases exponentially to represent the decay of viewer attention (or accumulation of boredom) and is "corrected" (reduced) when a stimulating event (a trailer shot) occurs.
$$\lambda(t) = \exp(\alpha t - \sum_{t_i < t} \beta)$$

\begin{figure}[t]
\centering
\includegraphics[width=0.8\linewidth, height=4cm]{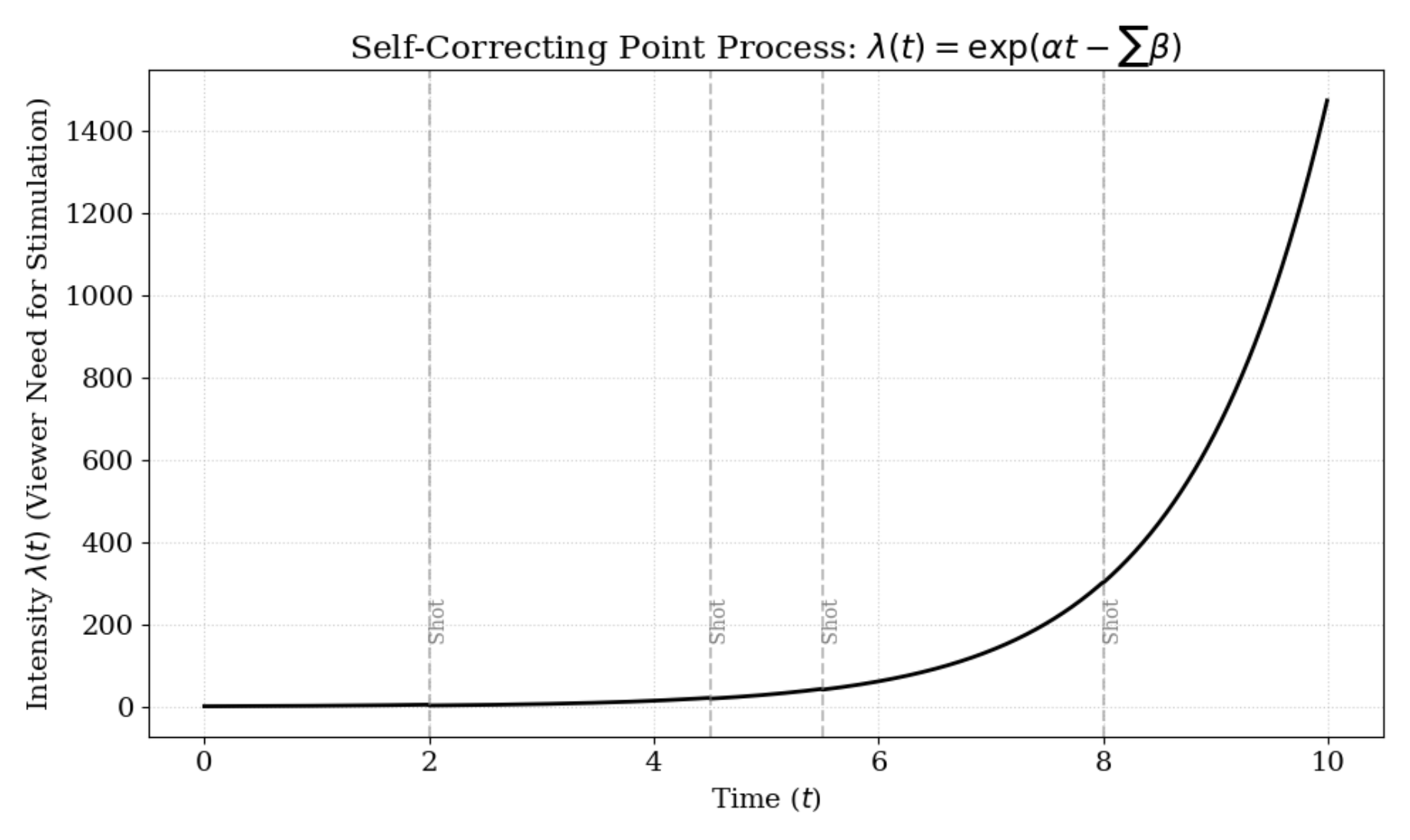}
\Description{A graph showing intensity function rising over time and dropping when an event occurs, illustrating the self-correcting point process.}
\caption{Self Correcting Point Process}
\label{fig:pp}
\end{figure}

This formulation captures the "rhythm" of a trailer, the necessity to intersperse high-intensity shots to maintain viewer engagement [37]. It moves beyond selecting individual good shots to modeling the sequence of shots required to sustain attention. Similarly, the Vid2Trailer framework (Irie et al., 2010) utilized affective content analysis, extracting "symbols" (logos, theme music) and ranking shots based on Bayesian surprise models to maximize emotional impact [17]. By modeling "surprise" as the Kullback-Leibler divergence between a viewer's prior and posterior beliefs about the video content, they could mathematically identify moments that defied expectation, a key component of effective trailers. Other works have explored intelligent facilitators for film production [4] and specific methodologies for generating teasers for long documentaries [25], further refining the extraction process.

\subsection{GCNs and Narrative Beats}

While affective models capture local saliency and immediate viewer reaction, they often fail to model the long-term narrative structure of a film. A trailer composed solely of explosions might be "attractive" in the short term but confusing and exhausting in the long term. To address this, Hu et al. (2022) proposed a \textbf{GCN-based framework} that treats a movie not as a linear stream, but as a topological graph where nodes are shots and edges represent semantic or temporal relationships [3].

Crucially, this approach integrated domain knowledge from screenwriting theory, specifically Blake Snyder's "Save the Cat" beat sheet structure. By performing \textbf{stratified sampling} based on 15 standard story beats (e.g., "Catalyst," "Fun and Games," "All Is Lost," "Finale"), the model ensures that the selected shots represent the structural arc of the film rather than just random high-energy moments. The Graph Convolutional Network aggregates information from neighbor shots to refine the feature representation, allowing the system to understand the contextual importance of a shot beyond its immediate visual pixels. For example, a quiet shot of a character reacting might be visually simple, but if it is graphically connected to a major plot point, the GCN can upweight its importance. This aligns with surveys on recent video summarization techniques [24], which emphasize the need for semantic preservation.

\section{LLM Orchestration}

The integration of Large Language Models (LLMs) marks the first phase of the generative era in trailer production. Rather than treating video as a narrative entity to be "read" and "rewritten" by an AI agent.

\subsection{The Multi-Stage Generative Pipeline}
A comprehensive framework introduced by Balestri et al. (2024) utilizes OpenAI's GPT-4 not merely as a text processor, but as a creative director [8]. This approach, often termed "LLM Orchestration," splits the trailer generation process into four different generative stages, utilizing the LLM reasoning capabilities to guide specialized sub-modules.

\textbf{(1) Preparation (Reasoning and Segmentation):} The process begins with the LLM analyzing synopsis and script of the video. The model is prompted to act as a video/film editor, segmenting the narrative into sub-plots and selecting key "visual" scenes that convey the theme. Crucially, the LLM is instructed to enforce anti-spoiler mechanisms, filtering out plot points that occur in the resolution phase of the video's narrative. This reasoning step is similar to the "paper edit" phase of human editing. \textbf{(2) Visual Retrieval (Semantic Matching):} Unlike previous methods, this framework employs CLIP (Contrastive Language-Image Pre-training). The LLM generates textual descriptions of the desired shots (e.g., A dusty farm under a fading sky, A tense conversation in a cockpit), and the system retrieves the frames from the movie that maximize the cosine similarity with these text prompts in the CLIP embedding space. This allows for semantic selection, finding specific concepts and imagery rather than just signal-based selection. \textbf{(3) Voice-Over Generation (Textual Synthesis):} The LLM generates a new, coherent voice-over script that is designed specifically for the trailer format (e.g., "In a world where time is running out..."). This script is not extracted from the movie dialogue but is synthesized \textit{de novo} to fill up the narrative gaps. Post that it is synthesized into audio using Text-to-Speech (TTS) models like Coqui xtts-v2, selecting voice profiles that match the movie's genre. \textbf{(4) Soundtrack Composition (Audio Synthesis):} The system uses the LLM to generate a musical prompt (e.g., "Ominous, slow tempo, cello and synthesizer, building to a crescendo"). This prompt is fed into a text-to-audio model like MusicGen or VidMuse [2, 15, 27] to create a unique background score that matches the emotional arc which is defined in the preparation phase.

\subsection{Hybrid Human-AI Co-Creation}
While fully automated pipelines are efficient, industrial applications often require a "Hybrid Human-AI" paradigm to ensure brand safety and narrative precision. Modern iterations of this approach are evident in educational and podcast domains. Mishra et al. (2022) describe a semi-automatic framework for learning pathways where the AI generates trailer fragments (e.g., course outlines) based on content, but enforces template constraints (e.g., using DSL approaches [39]) and creator input to prevent hallucinations in non-fiction contexts [6][7].

Similarly, the "PodReels" system allows human editors to refine AI-selected highlights for podcasts [22]. This creates a reinforcement learning loop: the AI suggests cuts based on transcript and audio analysis, and human acceptance/rejection signals are used to fine-tune the selection model. This hybrid model dominates professional settings where factual correctness and specific narrative controls take precedence over pure automation with a futuristic thinking of using facial expression for refining model [38].

\section{End-to-End Neural Synthesis}

While LLM orchestration relies on discrete modules (LLM + CLIP + TTS), recent research has focused on end-to-end differentiable architectures that can learn the "language" of trailers directly from data. These models aim to internalize the rules of editing, pacing, consistency, and juxtaposition within the neural network's weights.

\subsection{Trailer Generation Transformer (TGT)}

Argaw et al. (2024) proposed the \textbf{Trailer Generation Transformer (TGT)} that completely reformulates the problem, utilizing advanced multimodal data analysis [5] to treat generation as a Sequence-to-Sequence task [9].

\begin{figure}[t]
    \centering
    \includegraphics[width=0.8\linewidth]{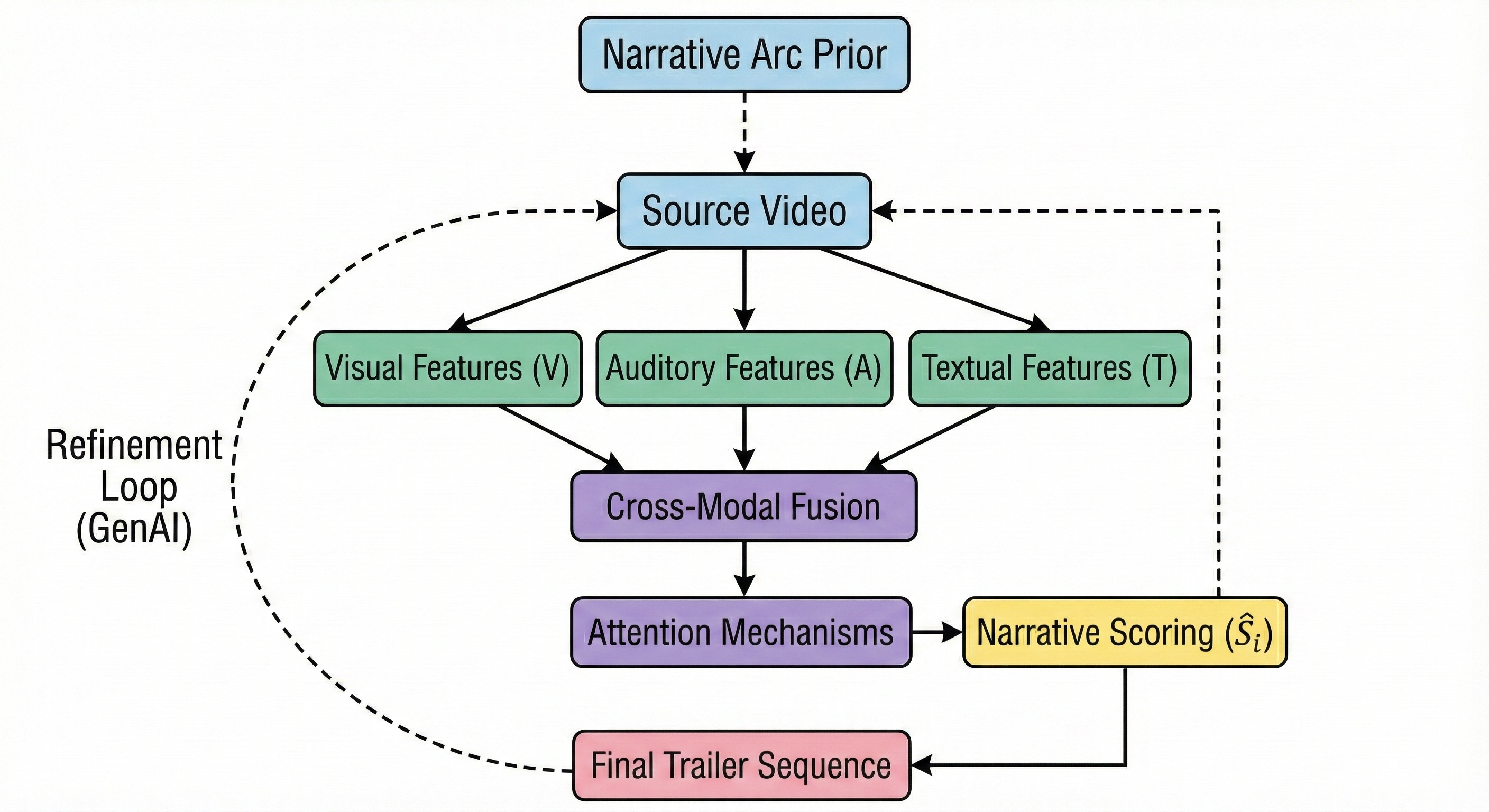}
    \Description{A block diagram of the Trailer Generation Transformer architecture showing the encoder and decoder flow.}
    \caption{Trailer Synthesis}
    \label{fig:trailer_synthesis}
\end{figure}
The architecture comprises new components designed to handle the specific dynamics of video:

\textbf{(1) Trailerness Encoder:} This module scans the entire movie sequence and predicts a "trailerness score" for each shot. These scores are not used to immediately select shots, but instead, they are injected into the visual embeddings as a form of continuous positional/attribute encoding. This allows the subsequent layers to "see" the global distribution of exciting moments before making local decisions.

\textbf{(2) Context Encoder:} This stack of Transformer encoder layers processes the movie sequence (enriched with trailerness scores) to create contextualized shot representations. Through self-attention, each shot's embedding is updated based on its relationship to every other shot in the movie. This allows the model to understand that a specific shot (e.g., a gun firing) relates to another shot (e.g., a character falling) even if those are far apart in the original source footage.

\textbf{(3) Autoregressive Trailer Decoder:} This is the core innovation. Unlike ranking methods that select the top-$K$ shots independently, TGT uses a Transformer decoder to predict the representation of feature for the next trailer shot that is conditioned on the previously generated shots. $$P(\hat{v}_j | C, v_1,..., v_{j-1})$$ Here, $\hat{v}_j$ is the predicted embedding for the $j$-th shot of the trailer, and $C$ is the contextualized movie representation. This autoregressive property is crucial because it enables the model to learn shot composition which is the artistic ordering of shots (e.g., establishing shot to action to reaction) that creates narrative flow, instead of simply outputting shots in chronological order.

\subsection{Performance and Benchmarking}

The TGT framework was evaluated on benchmarks derived from MovieNet, MAD, and the recently released Mmtrail dataset [21]. Quantitative results demonstrated significant improvements over previous state-of-the-art methods like CCANet and CLIP-It. TGT achieved an F1-score of 52.38\% on the MAD benchmark, significantly outperforming CLIP-It (41.73\%) and CCANet (31.63\%) [9]. More importantly, the Levenshtein Distance (LD) metric showed that TGT produces sequences much closer to professional edits (21.18 vs 95.58).

\section{The Frontier: Foundation Models}

The most recent development in this field is the emergence of general-purpose video generation foundation models.

\subsection{Sora and Veo 2: Physics and Fidelity}

Models like OpenAI's \textbf{Sora} [30, 31], Google's \textbf{Veo 2}, and Meta's Movie Gen [32] represent a quantum leap in generative capabilities [33]. These latent diffusion models can generate video content from textual prompts with temporal consistency. They also adhere to physical laws (e.g., reflections, fluid dynamics, object permanence) that were previously impossible.

In the context of trailer generation, these models offer capabilities that blur the line between editing and VFX:

\textbf{(1) Abstractive Fill-in:} If a movie lacks a specific connecting shot required for a trailer's narrative arc (e.g., an aerial view of a city to establish scale, or a close-up of a hand turning a doorknob), these models can generate a high-quality, style-consistent shot to fill the gap.

\textbf{(2) Style Transfer and Resynthesis:} Existing footage can be upscaled, restyled, or dynamically altered to fit the tone of a trailer. For instance, a daylight scene could be regenerated as a night scene to make it more ominous for a horror trailer, or the color grading could be shifted to match a specific aesthetic.

\subsection{Controllability vs. Creativity}

A major open challenge identified in recent surveys is \textbf{controllability}. While models like Sora generate visuals, forcing them to adhere to the precise narrative constraints of a specific source video remains difficult. Current research is pivoting towards \textbf{Video-to-Trailer Editing}, where the GenAI model takes a long source video as a constraint and acts as an intelligent refiner. The goal is to use the generative model to enhance extraction, generating smooth transitions between jump cuts, stabilizing shaky footage, or extending shots that are too short.

\section{Industrial Impact and Economics}

The transition to AI-driven trailer generation is reshaping the content production pipeline [34], moving from a manual, low-volume workflow to an automated, high-volume industrial process suitable for User-Generated Content (UGC) platforms.

\subsection{Content Velocity on Public Platforms}

Platforms like YouTube, TikTok, and Instagram Reels operate on the logic of \textbf{content velocity}, the speed at which new content is uploaded and consumed. For these platforms, the trailer (or "Short") is a content unit that drives the algorithmic discovery of long-form content. Automated trailer generation enables "Micro-Creators," who lack the resources for professional editors, to compete with professional studios. By reducing the editing time for a promotional "Short" or "Reel" from hours to seconds, AI tools lower the barrier to entry and increase the overall volume of high-engagement content entering the platform's recommendation engine [12].

\subsection{Scalability and Inference}

However, this comes with a computational cost. Recent economic modeling suggests that compute cost is the primary bottleneck for scaling these systems to the billions of videos on UGC platforms.

\textbf{Inference Load:} Processing 1 million hours of video monthly to generate trailers using heavy multimodal encoders (ViT-H, audio transformers) requires an estimated 15,000 to 20,000 GPU-hours on A100-class hardware.

\textbf{Optimization:} To make this viable for platforms hosting billions of videos, techniques like model distillation [38], quantization, and edge deployment are essential. The computational cost for massive deployment is estimated to be minimal per unit. While individual inference is computationally efficient, the aggregate cost becomes significant when scaling to billions of videos on UGC platforms, making optimization techniques like quantization essential. Efficient deployment architectures that process video at the edge or use tiered processing (lightweight models for initial selection, heavy models for final rendering) are areas of active research.

\subsection{The "Weak Trope" Mining}

Smith et al. (2017) introduced the concept of analyzing \textbf{Tropes}, storytelling devices like "Heroic Sacrifice," "Jump Scare," or "Meet Cute", to bridge the semantic gap [16]. By mining these tropes from massive datasets (MovieQA), AI systems can identify the visual elements (e.g., dark lighting and  zooming face make a horror trope) that correlate with high engagement for specific genres.

This data-driven approach allows industrial systems to tailor trailers to specific audience segments. A single movie could spawn dozens of micro-trailers: an action-heavy trailer for action fans, a romance-heavy trailer for drama fans, and a comedy-focused trailer for casual viewers. This \textbf{Personalized Trailer Generation} [40] represents the ultimate industrial application of the technology, optimizing the narrative hook for the specific viewer profile to maximize Click-Through Rate (CTR).

\section{Evaluation Metrics}

Evaluating generative trailers is notoriously difficult due to the inherent subjectivity of "creativity" and the multi-objective nature of the task. As the field transitions from extractive summarization to generative, the evaluation paradigm must also shift. Traditional metrics based on frame overlap are no longer sufficient when models can generate new frames and voice-overs, or radically reorder narrative structures. We propose a composite framework that assesses trailers across four dimensions: Fidelity, Narrative, Multimodal Synchronization, and Safety.

\subsection{Fidelity}

Historically, trailer generation has been evaluated as a retrieval task. Metrics like Precision, Recall, and F1-score quantify the overlap between the AI-selected shots and a human-edited ground truth trailer. For example, the TGT framework achieved a state-of-the-art F1-score of 52.38 on the MAD dataset. However, these metrics suffer from the "validity of alternatives" problem: a trailer that utilizes different but equally effective shots is penalized, failing to capture the creative variety inherent in editing.

To address the rise of generative fill-in models like Sora and Veo, evaluation must move beyond binary overlap to video consistency quality metrics like Fréchet Video Motion Distance, which can measure the realism and temporal coherence of synthesized "bridge shots" or style-transferred footage.

\subsection{Narrative}

A collection of high-saliency shots does not constitute a narrative. As noted in Section 3.1, simple affective models may select exciting moments that are disjointed or confusing. Narrative evaluation requires measuring the logical flow of the trailer.

\textbf{Beat Alignment Score:} Building on the graph-based work of Hu et al., evaluators should measure how well the generated sequence adheres to recognized storytelling beats (e.g., Setup, Catalyst, Climax) rather than random sampling.

\textbf{Semantic continuity:} With autoregressive Transformers predicting shot sequencing, we propose using Large Language Models (LLMs) as evaluators to score the causality between shots (e.g., Does Shot B logically follow the action in Shot A?).

\begin{table}[t]
\caption{Comparison of Generation Techniques}
\label{tab:evaluation_short}
\small
\renewcommand{\arraystretch}{0.9}
\setlength{\tabcolsep}{4pt}
\centering
\begin{tabularx}{\linewidth}{l l X}
\toprule
\textbf{Dimension} & \textbf{Metric} & \textbf{Best Technique \& Rationale} \\
\midrule
\textbf{Structural} & Levenshtein & \textbf{TGT (Transformer)}: Best at capturing shot ordering and editing grammar [9]. \\
\addlinespace[3pt]
\textbf{Visual} & FVD & \textbf{Foundation Models}: Superior pixel-level synthesis and fill-in capabilities [33]. \\
\addlinespace[3pt]
\textbf{Narrative} & Beat Align & \textbf{Graph/LLM}: Enforces logical segmentation and story beat adherence [3]. \\
\addlinespace[3pt]
\textbf{Multimodal} & AV-Sync & \textbf{LLM Agents}: Best at synthesizing semantically aligned audio/voice [8]. \\
\addlinespace[3pt]
\textbf{Safety} & Factuality & \textbf{Heuristic}: Zero hallucination risk by strictly using existing frames. \\
\bottomrule
\end{tabularx}
\end{table}

\subsection{Multimodal Synchronization}

Current research highlights a disparity between visual attractiveness and audio-visual coherence. In user studies utilizing Mean Opinion Scores (MOS), AI-generated trailers frequently score lower on "Appropriateness" compared to human editors [22][18]. This is often due to a lack of synchronization between the visual cuts and the musical beats or voice-over pacing. To evaluate the "LLM Orchestrated" pipeline, we propose specific multimodal metrics:

\textbf{AV-Sync Alignment:} Measuring the temporal offset between visual cut boundaries and audio onset peaks (beats) in the synthesized soundtrack.

\textbf{Textual Relevance (BERTScore):} For pipelines that generate original voice-overs, text-similarity metrics must ensure the synthesized script aligns thematically with the movie's actual synopsis, preventing tonal dissonance.

\subsection{Safety and Hallucination}

The integration of foundation models introduces the risk of "hallucination," particularly in sensitive domains like healthcare or non-fiction. The ability of models like Veo to "fill in" missing data  creates a potential for misinformation [33]. A trailer that implies a confrontation that never happens in the film, or synthesizes dialogue that was never spoken, crosses the line from promotion to deception.

\textbf{Factuality Score:} In non-fiction contexts, generated trailers must be evaluated on their adherence to the source truth. A high-quality trailer for a medical procedure must not visualize anatomical details that contradict the source video.

\textbf{Bias and Representative Sampling:} Since models trained on engagement data (CTR) may prioritize sensationalism, evaluation must include "Fairness Metrics" to ensure the trailer represents the film's actual content distribution rather than just its most violent or sexualized outliers [12].

\section{Conclusion and Roadmap}

The field of Video Trailer Generation has matured from a sub-discipline of signal processing into a frontier of Multimodal Generative AI. Future systems will not just find shots; they will fix shots, generate voice-overs, and compose music, behaving as full-stack creative agents. Narrative coherence can be solved by modeling video editing as a language, using Transformers to predict the "grammar" of cinema rather than just the content of frames [9]. Industrial systems will generate personalized trailers on-the-fly, optimizing the narrative hook for specific viewer profiles (e.g., "horror-focused" vs. "romance-focused" cuts) to maximize Click-Through Rates [19].

\subsection{Future Research Directions}

\textbf{(1) Narrative-Aware Transformers:}
    While current autoregressive models excel at shot sequencing, they often lack long-term causal reasoning. Future architectures must integrate graph-based ``story beat'' structures (as seen in GCN approaches) directly into the attention mechanism of Transformers [3]. This ensures the generated sequence adheres to dramatic arcs (Setup $\to$ Climax) rather than just visual similarity.

\textbf{(2) The Evaluation Crisis: Standardizing Subjectivity:}
    As the field abandons retrieval metrics like Precision and Recall, there is an urgent need for standardized \textit{Generative Evaluation Protocols}. Future research must develop composite metrics that weigh ``Visual Fidelity'' (FVD) against ``Narrative Consistency'' (Beat Alignment Scores) and ``Auditory Appropriateness,'' potentially utilizing LLMs as semantic judges to score logical flow [13].

\textbf{(3) Semantic Consistency and Truthfulness:}
    The ability of foundation models to ``hallucinate'' plausible footage introduces a ``Truth-in-Advertising'' challenge. In domains ranging from entertainment to healthcare, systems must implement \textit{Consistency Alignment} constraints to ensure that synthesized ``fill-in'' shots or stylized edits remain factually representative of the source content, preventing the generation of misleading sensational shots [33][29].

\textbf{(4) Economic Viability and Inference Optimization:}
    Democratizing high-end trailer synthesis for UGC platforms requires solving the compute bottleneck. Research must prioritize model distillation and edge-deployment strategies to reduce the estimated 15,000+ GPU-hours required for mass-scale processing, ensuring that personalized generation is economically sustainable [14].

\textbf{(5) Provenance and Governance:}
    As the distinction between human-edited and AI-generated content vanishes, implementing robust provenance tracking (e.g., C2PA) is essential which must track not only the origin of shots but the \textit{intent} of the edit, preserving human artistic oversight in an  algorithmic loop [22].

\section*{Acknowledgements}
The authors acknowledge the use of Google's Gemini AI to refine the text and generate the images included in this survey.

\let\oldbibliography\thebibliography
\renewcommand{\thebibliography}[1]{%
  \oldbibliography{#1}%
  \setlength{\itemsep}{1pt}%
  \setlength{\parskip}{0pt}%
  \small
}

\end{document}